%% file: main.tex
\documentclass{WileyMSP-template}

\usepackage[utf8]{inputenc}
\usepackage[T1]{fontenc}
\usepackage{hyperref}
\usepackage{url}
\usepackage{booktabs}
\usepackage{multirow}
\usepackage{subcaption}
\usepackage{amsfonts}
\usepackage{amsmath}
\usepackage{amssymb}
\usepackage{nicefrac}
\usepackage{microtype}
\usepackage{xcolor}
\usepackage{graphicx}
\usepackage{float}
\usepackage{placeins}
\usepackage{siunitx}
\usepackage{listings}
\usepackage{bm}
\usepackage[ruled,vlined]{algorithm2e}
\usepackage{tikz}
\usetikzlibrary{positioning,arrows.meta}
\usepackage{dsfont}

\newcommand{\vtau}{\boldsymbol{\tau}}

\DeclareSIUnit{\meter}{m}
\graphicspath{{figures/}}

\input{macros}

\makeatletter
\renewcommand{\maketitle}{%
  \begin{center}
    {\huge \bfseries \@title\par}\vspace{0.8em}
    {\large \AuthorList}\par
  \end{center}
  \vspace{0.8em}
}
\makeatother

\title{Co-Design of Rover Wheels and Control using Bayesian Optimization and Rover-Terrain Simulations}
\newcommand{\AuthorList}{Huzaifa Mustafa Unjhawala, Khizar Shaikh, Luning Bakke, Radu Serban, Dan Negrut*}

\begin{document}

\maketitle

\dedication{}

\begin{affiliations}
Department of Mechanical Engineering, University of Wisconsin--Madison, Madison, WI 53706, USA\\
*Corresponding author: Dan Negrut (\texttt{negrut@wisc.edu})
\end{affiliations}

\keywords{Bayesian optimization, rover wheel design, continuum terramechanics, closed-loop simulation, co-design, SPH, deformable terrain}

\begin{abstract}
While simulation is vital for optimizing robotic systems, the high cost of modeling deformable terrain has historically prohibited its use in full-vehicle design studies that involve off-road autonomous mobility. For instance, Discrete Element Method (DEM) simulations are often restricted to single-wheel tests, limiting conclusions about integrated wheel-vehicle-controller interactions and hindering joint optimization of system design and system control. This paper presents a Bayesian optimization framework that co-designs rover wheel geometry and steering controller parameters using high-fidelity, full-vehicle closed-loop simulations on deformable terrain. Leveraging the computational efficiency and scalability of a continuum-representation model (CRM) for terramechanics, we evaluate candidate designs on trajectories of varying complexity while towing a fixed load. The optimizer tunes wheel parameters (radius, width, and grouser features) and steering PID gains under a multi-objective formulation that balances traversal speed, trajectory tracking error, and energy consumption. We compare two optimization strategies -- simultaneous co-optimization of wheel and controller parameters versus a sequential approach that decouples mechanical and control design -- and analyze their trade-offs in terms of performance and computational cost. Covering 3,000 full-vehicle simulations, these optimization campaigns conclude within five to nine days -- a process that formerly required months when using the group's previous DEM-based approaches. Finally, we report a preliminary hardware study indicating that the simulation-optimized wheel designs preserve relative performance trends on the physical rover. Together, these results show that scalable high-fidelity simulation can enable practical co-optimization of wheel design and control for off-road vehicles on deformable terrain without relying on computationally prohibitive DEM studies. The simulation infrastructure (scripts and models) used in this study is available as open source in a public repository for reproducibility studies and further research.
\end{abstract}

\input{sections/intro}
\input{sections/method}

\input{sections/results}
\input{sections/conclusion}
\input{sections/future}
\FloatBarrier

\bibliographystyle{MSP}
\bibliography{references}

\end{document}

%% file: macros.tex
\definecolor{arsenic}{rgb}{0.23, 0.27, 0.29}
\definecolor{charcoal}{rgb}{0.21, 0.27, 0.31}
\definecolor{hanblue}{rgb}{0.27, 0.42, 0.81}
\definecolor{blue-ncs}{rgb}{0.0, 0.53, 0.74}
\definecolor{awesome}{rgb}{1.0, 0.13,0.32}
\definecolor{darkgreen}{rgb}{0, .4,0}

\newcommand{\be}{\begin{equation}}
\newcommand{\ee}{\end{equation}}
\newcommand{\bd}{\begin{displaymath}}
\newcommand{\ed}{\end{displaymath}}
\newcommand{\BE}{\begin{eqnarray}}
\newcommand{\EE}{\end{eqnarray}}

%% file: sections/intro.tex
\section{Introduction}
\label{sec:intro}
Simulation enables ``virtual'' experiments with low cost and low risk and has become an essential tool in the engineering of robotic systems~\cite{PNASsimRobotics2021}. In the context of mobile robots and rovers, simulation allows engineers to evaluate control strategies, explore failure modes, and assess performance in challenging environments before risking damage to physical hardware. This capability is particularly important for off-road and planetary vehicles, where wheel--terrain interactions are highly nonlinear and difficult to study experimentally. In such settings, simulation provides a systematic means to investigate physical phenomena that are impractical or misleading to reproduce in laboratory tests. For example, a recent study demonstrated that ``gravity offset'' experiments can significantly overestimate rover tractability, arguing instead for simulation-based evaluation of mobility performance~\cite{gravOffset2025}. As a result, physics-based simulation has become standard for control development, algorithm benchmarking, and system validation in robotics~\cite{PNASsimRobotics2021,negrut2021PhysicsSimulators}.

Despite its widespread adoption for control and validation, simulation has seen comparatively limited use as a tool for rover wheel design, particularly when deformable terrain is involved. This is largely due to the high computational cost of accurately modeling granular media, which has constrained most simulation-based wheel optimization studies to simplified, single-wheel setups.

\subsection{Relevant Work}
\label{sec:relevant_work}
One of the few notable examples of simulation-driven rover wheel design is the work of Stubbig and Lichtenheldt at DLR~\cite{DLR2021RoverWheelDesign}, who applied Bayesian Optimization (BO) to single-wheel Discrete Element Method (DEM) simulations for the MMX (Phobos) rover mission. In their study, wheel geometry was parameterized using four variables, i.e., grouser height, number, chevron angle, and rim curvature, and each candidate design was evaluated under three driving scenarios: forward motion, reverse motion, and uphill traversal on a $10^\circ$ slope. Performance was measured using the total traveled distance normalized against a baseline wheel design. The resulting optimized wheel achieved an approximately $64\%$ improvement in simulated distance relative to the prior prototype. Owing to the high computational cost of DEM, however, the optimization was restricted to a single-wheel test with free-slip, fixed-angular-velocity conditions. The wheel was constrained to remain upright with no lateral motion, precluding steering effects and coupling with other wheels to capture load transfer effects. The BO process comprised 115 iterations, with each simulation requiring approximately 20 hours on an Nvidia V100 GPU, resulting in a total compute time of roughly 96 days.

A similar approach was adopted in~\cite{ruochunPhDthesis2023} for the MGRU3 rover, where Bayesian Optimization was used to improve wheel geometry using DEM-based simulations. This study employed the DEME solver~\cite{ruochun2024dem} and focused on single-wheel performance metrics such as slope traction and a steering-related quantity defined as $S_t = \frac{F_y}{mg}$, representing lateral force per unit weight when the wheel is misaligned by $5^\circ$. As in the DLR study, the wheel was constrained to forward motion without full planar freedom, and angular velocity was externally prescribed rather than governed by a controller. Approximately 1744 simulations were evaluated, each taking around 30 minutes on an Nvidia A100 GPU, for a total compute time of roughly 36 days.

\subsection{Gaps in the Literature}
\label{sec:gaps_in_literature}
Together, these studies demonstrate that modern DEM-based simulation enables systematic exploration of rover wheel design spaces beyond what is feasible with purely empirical methods. However, they also highlight several persistent limitations. By relying on isolated single-wheel tests, these approaches neglect vehicle-level dynamics, wheel--wheel coupling, and the interaction between wheel geometry and closed-loop control. Fixed constraints on motion and externally imposed actuation further limit their ability to capture realistic maneuvering behavior such as steering. Moreover, despite advances in GPU-accelerated DEM solvers, the computational cost of these studies remains substantial, with optimization campaigns requiring over a month of wall-clock time.

\subsection{Contribution}
\label{sec:contribution}
We introduce a full-vehicle, closed-loop optimization framework for rover wheel design and control, using the continuum-based terramechanics simulator Chrono::CRM~\cite{Huzaifa2025CRM,weiGranularSPH2021} as the simulation backend. The framework embeds wheel geometry, steering controller gains, and vehicle dynamics within a BO loop, allowing vehicle-level evaluation of traction, steering behavior, and energy efficiency under closed-loop operation. The contributions of this work are as follows:
\begin{itemize}
    \item We present an optimization framework that simultaneously tunes rover wheel geometry (radius, width, grouser features, and orientation) and steering PID gains within full-vehicle, closed-loop simulations on deformable terrain.
    \item We formulate a multi-objective cost function that balances maneuver completion time, trajectory tracking error, and energy consumption.
    \item We compare two optimization strategies, i.e., simultaneous co-optimization of wheel and controller parameters versus a sequential approach that decouples mechanical and control design, and analyze their trade-offs.
    \item We report a preliminary experimental validation using a payload pull test, showing that relative performance rankings from simulation are preserved on the physical rover.
    \item We complete optimization campaigns of 3,000 full-vehicle simulations in five to nine days depending on strategy, compared to the month or more required by prior DEM-based studies.
\end{itemize}

\subsection{Organization}
\label{subsec:organization}
Section~\ref{sec:methodology} describes the vehicle and terrain models, wheel parameterization, and BO setup. Section~\ref{sec:results} presents optimization results, sensitivity analyses, and experimental validation. Section~\ref{sec:conclusion} summarizes the findings, and Section~\ref{sec:future} discusses future work. Code and instructions to reproduce the results are available at \url{https://github.com/uwsbel/sbel-reproducibility/tree/master/2025/WheelOpt}.

%% file: sections/method.tex
\section{Methodology}
\label{sec:methodology}
\begin{figure}[H]
    \centering
    \includegraphics[width=0.4\textwidth]{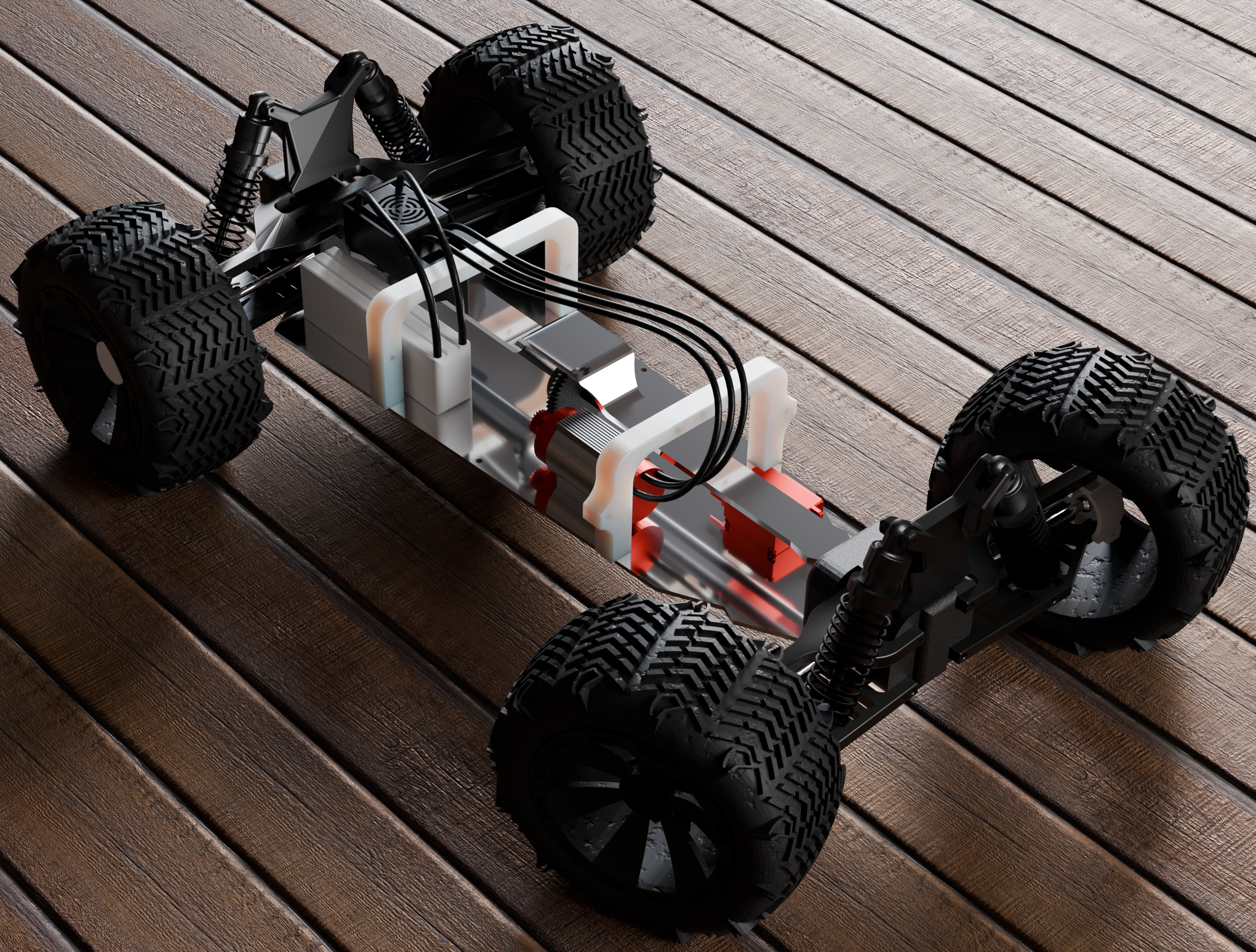}
    \caption{The \texttt{ART} vehicle used in this study. The platform has a wheelbase of \SI{0.47}{m} and a track width of \SI{0.34}{m}.}
    \label{fig:art_render}
\end{figure}

This section describes the simulation framework and optimization approach used to simultaneously co-design the wheel geometry and steering control policy for the rover. All experiments, both physical and virtual, were conducted using the Autonomy Research Testbed (\texttt{ART})~\cite{artatkResearchPlatform2022,ishaanGPSsim2realGap2024}, a 1/6th-scale rover platform with a wheelbase of \SI{0.47}{\meter} and a track width of \SI{0.34}{\meter}, see Fig.~\ref{fig:art_render}. This scale is comparable to that of the MMX rover~\cite{MMX2020Sedlmayr}. The vehicle is simulated using Chrono::Vehicle~\cite{chronoVehicle2019}, coupled with deformable terrain modeled using Chrono::CRM~\cite{Huzaifa2025CRM}; details are provided in Sec.~\ref{subsec:simulation}.

Bayesian optimization of wheel geometry and steering PID gains is performed using the Adaptive Experimentation Platform (Ax)~\cite{olson2025ax} with a BoTorch backend~\cite{balandat2020botorch}. The overall framework is summarized in Sec.~\ref{subsec:overall_framework}, with the optimization setup detailed in Sec.~\ref{subsec:bayesian_optimization}.

\subsection{Vehicle and Terrain Modeling}
\label{subsec:simulation}
\subsubsection{\texttt{ART} with Chrono::Vehicle}
\texttt{ART} features double-wishbone independent suspension at both axles, a Pitman-arm steering mechanism, and an all-wheel-drive configuration with a center differential.

The vehicle is simulated using a digital twin, referred to as \texttt{dART}, implemented in Chrono::Vehicle, a module of the Project Chrono framework~\cite{chronoOverview2016} that provides physics-based modeling of vehicle dynamics, including powertrains, suspensions, steering systems, and driver controls. The \texttt{dART} model matches the physical platform in suspension geometry, steering mechanism, and drivetrain architecture, and has been experimentally calibrated and validated in prior work~\cite{ishaanGPSsim2realGap2024,elmquist2022art,huzaifaIEEE-AccessCalibration2024}.

\subsubsection{Terrain Modeling with Chrono::CRM}
Chrono::CRM models discrete granular material as a continuum, posing wheel--soil interaction as a fluid--structure interaction (FSI) problem. This continuum formulation captures the bulk mechanical response of granular media while enabling computational efficiency that would be prohibitive with discrete element methods. Here we outline the mathematical formulation; full details and validation are provided in~\cite{Huzaifa2025CRM}.

The dynamics of the granular medium is governed by conservation of mass and momentum. In Lagrangian form, the density $\rho$ and velocity $\mathbf{u}$ evolve according to~\cite{Huzaifa2025CRM}:
\begin{align}
    \label{eq:CRM_const}
    \frac{d\rho}{dt} &= -\rho \nabla \cdot \mathbf{u},\\
    \label{eq:CRM_eq}
    \frac{d\mathbf{u}}{dt} &= \frac{\nabla \cdot \boldsymbol{\sigma}}{\rho} + \mathbf{f}_b,
\end{align}
where $\boldsymbol{\sigma}$ is the Cauchy stress tensor and $\mathbf{f}_b$ denotes body forces per unit mass. Since the stress tensor is not determined by conservation laws alone, a constitutive relation is required for closure.

To ensure frame-indifference under rigid-body rotation, stress evolution is expressed using the Jaumann objective rate:
\begin{equation}
    \label{eq:jaumann_rate}
    \overset{\vartriangle}{\boldsymbol{\sigma}} = \frac{d\boldsymbol{\sigma}}{dt} - \dot{\boldsymbol{\phi}} \cdot \boldsymbol{\sigma} + \boldsymbol{\sigma} \cdot \dot{\boldsymbol{\phi}},
\end{equation}
where $\dot{\boldsymbol{\phi}} = \frac{1}{2}(\nabla \mathbf{u} - (\nabla \mathbf{u})^\mathrm{T})$ is the rotation-rate tensor. The Jaumann rate is related to the strain rate through a linear hypoelastic relation involving bulk modulus $K$ and shear modulus $G$. To capture irreversible deformation in granular media, the formulation adopts a small-strain rate elasto-plastic decomposition:
\begin{equation}\label{equ:strain_rate}
    \dot{\boldsymbol{\varepsilon}} = 
    \underbrace{\frac{1}{2}(\nabla\mathbf{u} + \nabla\mathbf{u}^\intercal)}_{\text{total strain rate}} 
    - 
    \underbrace{\frac{1}{\sqrt{2}}\dot{\lambda} \frac{\boldsymbol{\tau}}{\bar{\tau}}}_{\text{plastic strain rate}},
\end{equation}
where the plastic flow is governed by a $\mu(I)$-based rheology~\cite{dunatunga2017continuum} that models the rate-dependent frictional behavior characteristic of dense granular flow. Here, $\bar{\tau} = \sqrt{\frac{1}{2}\vtau:\vtau}$ is the equivalent shear stress and $\dot{\lambda}$ is the plastic shear strain rate.

The governing equations are discretized spatially using smoothed particle hydrodynamics (SPH). Each particle $i$ carries density $\rho_i$, velocity $\mathbf{u}_i$, and stress $\boldsymbol{\sigma}_i$, which evolve according to discrete analogs of Eqs.~\eqref{eq:CRM_const}--\eqref{equ:strain_rate} expressed as sums over neighboring particles weighted by an SPH kernel $W_{ij}$ and its gradient. The resulting semi-discrete system is integrated in time using a two-stage Runge--Kutta (RK2) scheme. Chrono::CRM executes these computations in parallel on the GPU, assigning one thread per particle. Simulating full-vehicle terrain interaction at near real-time rates is essential for the present work, where thousands of closed-loop simulations are required for optimization. Further implementation details are provided in~\cite{Huzaifa2025CRM}.

\subsubsection{Modeling the Vehicle--Terrain Interaction}
The vehicle multibody system and Chrono::CRM terrain model are coupled through an explicit force--displacement co-simulation framework implemented via Chrono::FSI~\cite{RaduFSI2025}. Chrono::FSI manages data exchange between the multibody vehicle dynamics and the deformable terrain simulation, enabling two-way interaction between the solid and granular phases. Specifically, the vehicle system communicates full state information at the position and velocity levels, while the CRM terrain model returns the resultant contact forces and torques acting on vehicle components in contact with the soil.

\begin{figure}[H]
    \centering
    \includegraphics[width=0.5\textwidth]{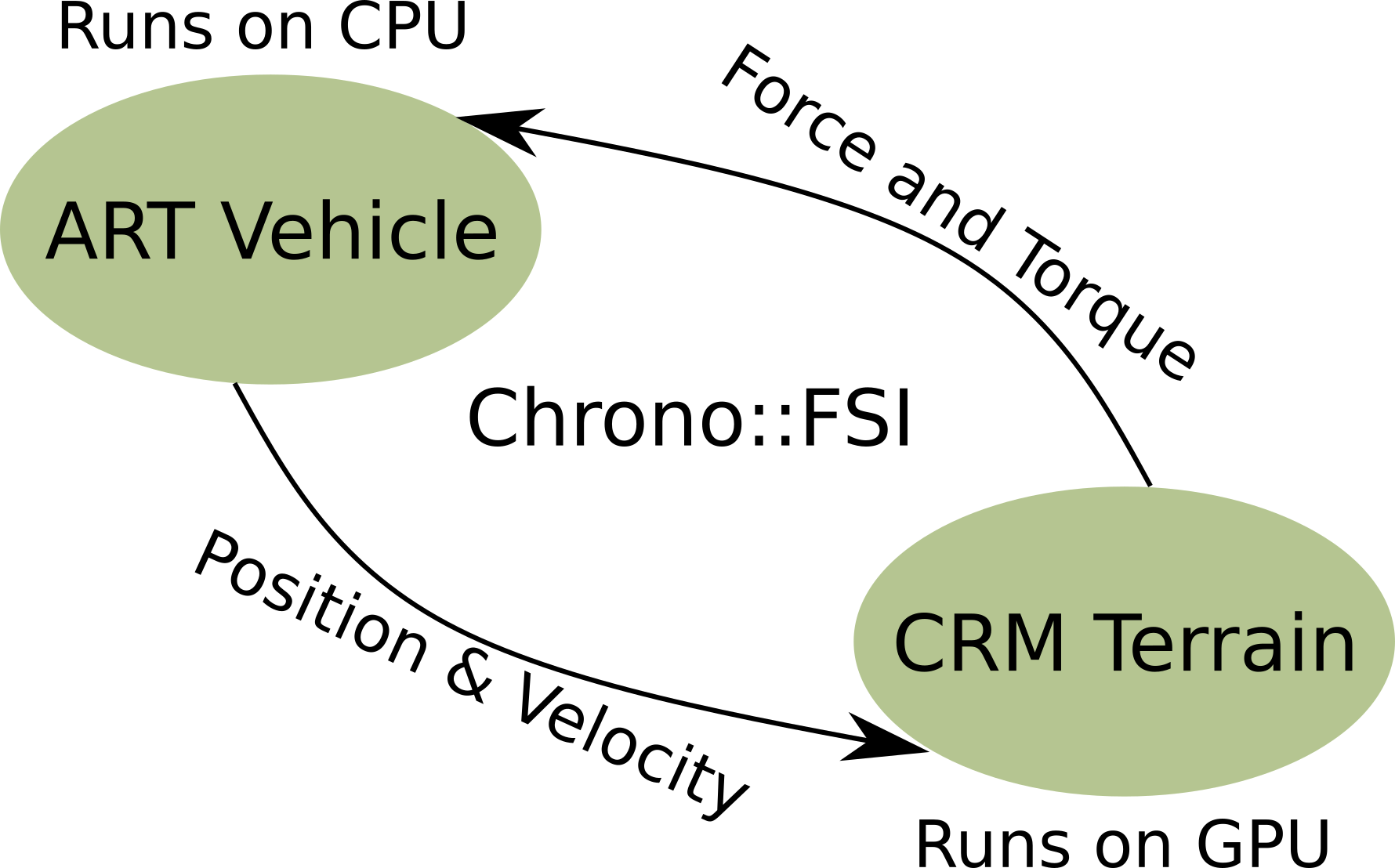}
    \caption{Schematic of the co-simulation framework. The vehicle multibody system exchanges state and force information with the SPH-based terrain at each time-step, with both subsystems advancing in parallel. CRM receives from ART location of the wheels; ART receives from CRM forces acting on the wheels.}
    \label{fig:cosim_schematic}
\end{figure}

In addition to handling data exchange, Chrono::FSI controls the co-simulation meta-step, which defines the frequency of synchronization between the two subsystems. The vehicle and terrain dynamics are advanced concurrently using non-blocking parallel threads, allowing both phases to evolve simultaneously. A schematic overview of the co-simulation framework is shown in Fig.~\ref{fig:cosim_schematic}. In this study, the meta-step is set equal to the time step of the vehicle multibody system.

Interaction between the particle-based CRM terrain and the solid bodies of the vehicle is enforced through Boundary Condition Enforcing (BCE) markers~\cite{Huzaifa2025CRM}. These markers are distributed over solid surfaces in multiple layers with an initial uniform spacing $d_0$, matching the spacing of the SPH particles used to represent the terrain. As the vehicle evolves in time, the BCE markers move rigidly with the associated solid body and act as specialized SPH particles whose kinematics are prescribed by the solid motion.

Although the position of each BCE marker is fully determined by the solid body they are attached to, the SPH marker also carries state variables such as velocity and stress, which are obtained through extrapolation from the surrounding soil particles. Chrono::CRM supports multiple extrapolation schemes for this purpose, including the \textit{Adami} and \textit{Holmes} methods~\cite{Huzaifa2025CRM}. In this work, the \textit{Adami} method is used for all simulations. The extrapolated state information is only employed to compute contact forces exerted by the soil on the solid body and to impose a no-slip boundary condition at the vehicle--terrain interface. Additional implementation details are provided in~\cite{Huzaifa2025CRM}.

\begin{figure}[H]
    \centering
    \includegraphics[width=0.8\textwidth]{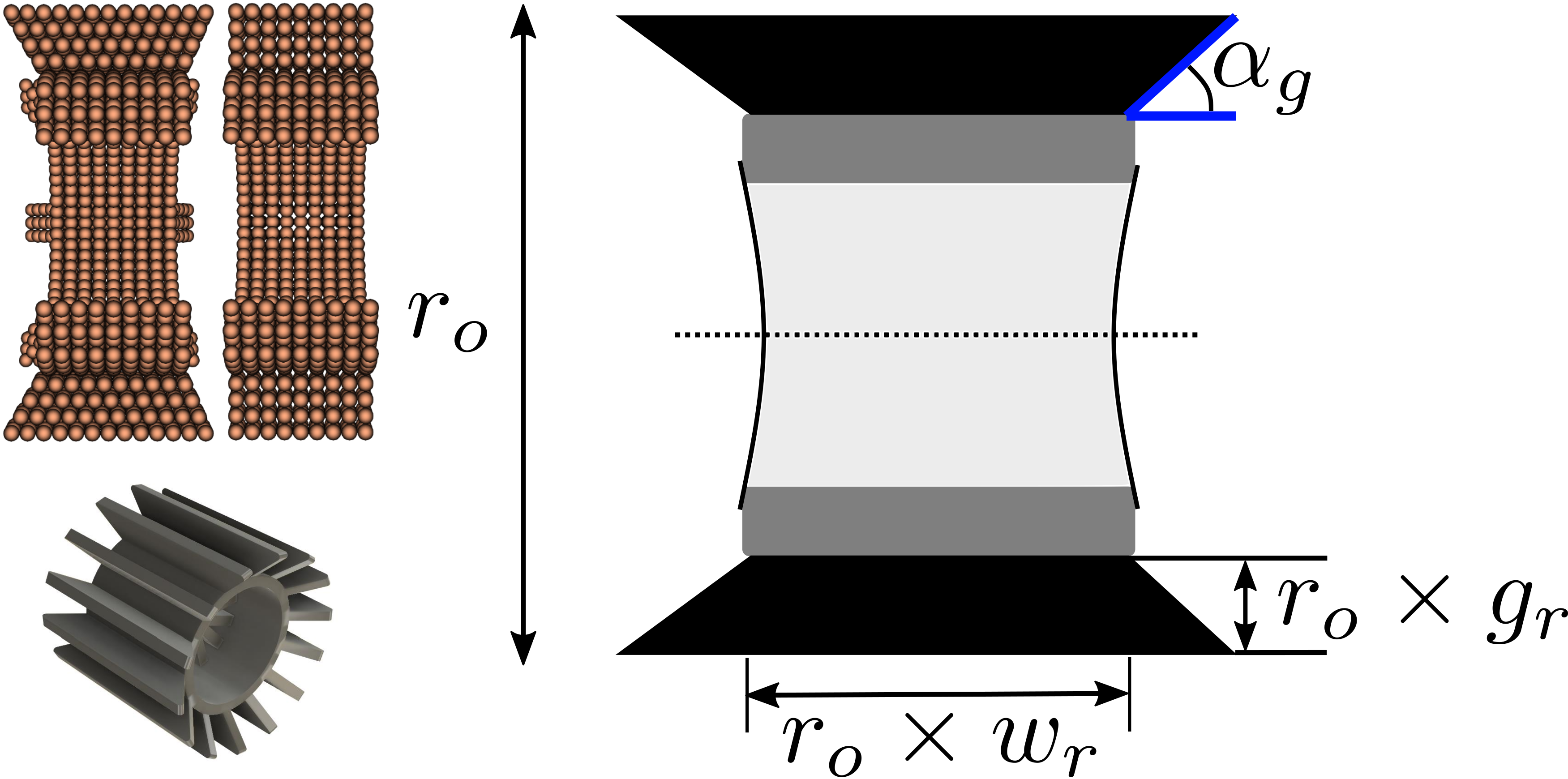}
    \caption{Left and top: BCE marker representations of two wheel designs generated during optimization. Left and Bottom: Isometric view of a CAD model of the wheel. Right: the geometric parameterization used for BO (see Sec.~\ref{subsec:bayesian_optimization}).}
    \label{fig:wheel_geometries}
\end{figure}

In the context of this paper, the primary solid bodies interacting with the terrain are the wheels of \texttt{dART}. This interaction model enables wheel geometry to be represented directly as a point cloud of BCE markers, uniformly spaced with spacing $d_0$ in all directions. Such a representation is particularly well suited for design optimization, as it allows arbitrary wheel geometries generated by the BO framework to be evaluated consistently within the same vehicle--terrain interaction model. Examples of wheel geometries produced by the optimization process are shown in Fig.~\ref{fig:wheel_geometries}.

Chrono::CRM has been validated against experimental benchmarks relevant to rover mobility, including traction and slip studies under reduced-gravity conditions~\cite{gravOffset2025,Schepelmann2025TireGroundModeling}, as well as sphere cratering and cone penetration tests~\cite{Huzaifa2025CRM}. These validations demonstrate that the continuum formulation captures key terramechanical behaviors, e.g., drawbar pull, slip-sinkage coupling, and grouser-induced shear failure, governing rover performance on deformable terrain. While DEM explicitly resolves particle-level interactions, it incurs computational costs that preclude full-vehicle, closed-loop optimization at the scale considered here. The continuum approach trades microscale granular detail for macroscopic fidelity at substantially lower cost; for design optimization, the salient point is that CRM correctly ranks candidate designs, a property we verify experimentally in Sec.~\ref{subsec:wheel_only}.

Equally important for this work, recent algorithmic and GPU-level optimizations allow Chrono::CRM to simulate full-scale vehicles on deformable terrain at near real-time speeds~\cite{Huzaifa2025CRM}. This combination of physical fidelity and computational scalability makes Chrono::CRM well suited for the present optimization study, where thousands of closed-loop simulations are required.

\subsection{Bayesian Optimization Framework}
\label{subsec:bayesian_optimization}

\subsubsection{Parameterization}
The wheel geometry is parameterized using five variables: outer radius ($r_o$), width ratio ($w_r$), grouser ratio ($g_r$), number of grousers ($n_g$), and grouser angle ($\alpha_g$). The outer radius of the wheel is the radius of the wheel including the grousers. The width ratio is defined as the ratio of wheel width to outer radius, while the grouser ratio specifies the ratio of grouser height to outer radius. The grouser angle controls the orientation of the grousers relative to the wheel circumference. Design bounds are imposed based on vehicle geometry and manufacturing constraints: $r_o \in [0.05, 0.14]~\si{m}$, $w_r \in [0.7, 1.4]$, $g_r \in [0.0, 0.3]$, $n_g \in [2, 16]$, and $\alpha_g \in [45, 135]^\circ$.

This parameterization was selected to balance expressiveness and manufacturability. Compared to more complex parameterization used in prior work~\cite{ruochunPhDthesis2023,DLR2021RoverWheelDesign}, the chosen design space can express diverse wheel geometries while remaining practical to fabricate. In particular, allowing the grousers to fan outward through $\alpha_g$ increases the effective contact width without increasing the base wheel width.

In addition to wheel geometry, steering control is implemented using a PID controller with gains $K_{p,s} \in [0, 15]$, $K_{i,s} \in [0, 5]$, and $K_{d,s} \in [0, 5]$. Only the steering gains are included in the optimization, while throttle PID gains are held fixed. The target forward velocity is set sufficiently high ($3\,\si{m/s}$) that the throttle controller operates near saturation on deformable terrain, effectively commanding maximum available torque regardless of gain values. Under these conditions, throttle gains have negligible influence on traversal performance, which is instead limited by wheel--terrain traction. This choice also reduces the dimensionality of the design space, improving sample efficiency without sacrificing performance. The complete design vector is therefore defined as
\begin{equation}
\label{equ:design_parameters}
\mathbf{x} = [r_o, w_r, g_r, n_g, \alpha_g, K_{p,s}, K_{i,s}, K_{d,s}]^\top \in \mathbb{R}^8.
\end{equation}

\subsubsection{Proposed Framework}
Optimizing the design vector $\mathbf{x}$ requires repeated evaluation of a computationally expensive, black-box objective defined implicitly through high-fidelity, closed-loop vehicle--terrain simulation. To efficiently explore this design space under a limited simulation budget, we employ a sample-efficient, gradient-free BO optimization framework.

Let $f(\mathbf{x}) = \mathcal{J}(\mathcal{F}(\mathbf{x}))$ denote the scalar objective function, where $\mathcal{F}$ represents the Chrono simulation and $\mathcal{J}$ maps simulation outcomes to a performance metric. The objective function is designed to jointly capture traversal speed, path-tracking accuracy, and energy consumption during the maneuver.

\paragraph{Objective Function:}
For a given design $\mathbf{x}$, the simulator produces a trajectory from which a single, dimensionless composite cost is computed as
\begin{equation}
    \mathcal{J}(\mathcal{F}(\mathbf{x})) = w_s\, r_t + w_t\, e_n + w_p\, p_n,
\end{equation}
where $w_s$, $w_t$, and $w_p$ are non-negative weights satisfying $w_s + w_t + w_p = 1$. Lower values of $\mathcal{J}$ indicate better performance. Each term is normalized by a reference value and scaled by a common factor of 10 to produce comparable magnitudes across the three objectives.

The time term $r_t$ penalizes slow traversal relative to an ideal completion time based on a target speed of $v_{\text{target}} = 5\,\si{m/s}$:
\[
    r_t = 10\,\frac{t_{\text{elapsed}}}{L_{\text{path}} / v_{\text{target}}}.
\]
The tracking term $e_n$ captures path-following accuracy using the RMS cross-track error $e_{\text{rms}}$ in the horizontal plane, computed as the nearest distance to a polyline interpolation of the path defined by Bézier waypoints. This error is normalized by an ideal reference value $e_{\text{ideal}} = 0.01\,\si{m}$,
\[
    e_n = 10\,\frac{e_{\text{rms}}}{e_{\text{ideal}}}.
\]
Energy efficiency is quantified through the average mechanical power
$P_{\text{avg}} = \overline{\tau_m\,\omega_m}$, computed after an initial $0.5\,\si{s}$ transient and normalized by a reference power $P_{\text{ideal}} = 20\,\si{W}$,
\[
    p_n = 10\,\frac{P_{\text{avg}}}{P_{\text{ideal}}}.
\]

\paragraph{Surrogate Model, Initialization, and Acquisition:}
The objective function $f(\mathbf{x})$ is modeled using a Gaussian Process surrogate with a scaled Matérn-$5/2$ kernel, as implemented in BoTorch~\cite{balandat2020botorch}. The integer-valued parameter $n_g$ is treated as continuous during optimization and rounded to the nearest integer for simulation evaluation. The optimization process begins with an initial space-filling design generated using a Sobol sequence to ensure broad coverage of the design space and a well-conditioned kernel matrix. After this initialization phase, candidate designs are selected sequentially by maximizing the Log Noisy Expected Improvement (LogNoisyEI) acquisition function, which balances exploration and exploitation while improving numerical stability in the presence of noise.

At each iteration, the selected design vector is evaluated using the full closed-loop Chrono::CRM simulation, and the resulting objective value is incorporated into the GP surrogate. This process is repeated until the allocated simulation budget is exhausted. The choice of objective weights is discussed in Sec.~\ref{sec:results}.
\subsection{Overall Framework}
\label{subsec:overall_framework}
\begin{figure}[H]
    \centering
    \includegraphics[width=0.8\textwidth]{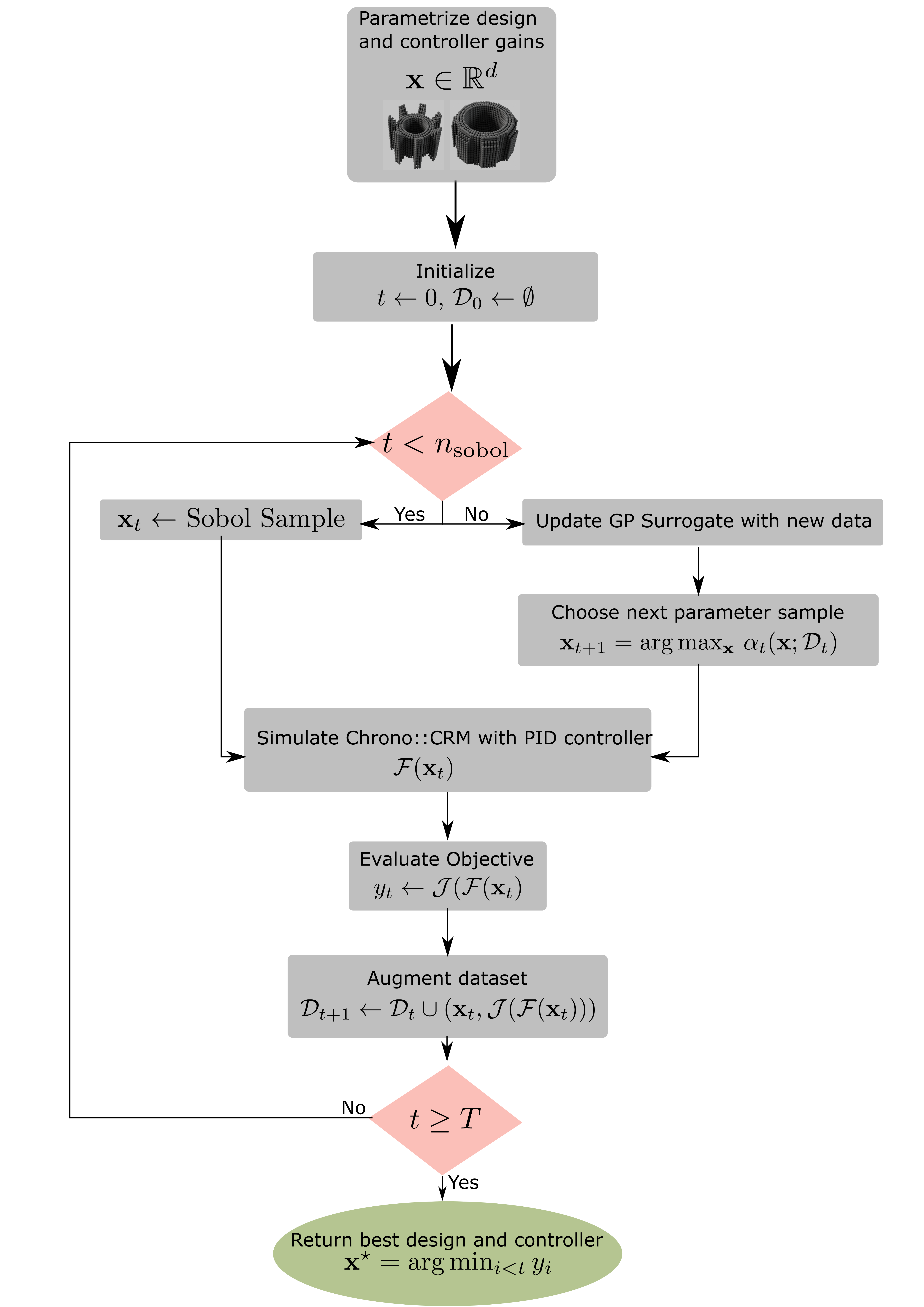}
    \caption{Optimization workflow. The loop alternates between Sobol sampling (during initialization) and acquisition-based selection, with each candidate evaluated via Chrono::CRM simulation. The process terminates after $T$ total evaluations, returning the design with the lowest objective value.}
    \label{fig:bo_flowchart}
\end{figure}

Figure~\ref{fig:bo_flowchart} summarizes the complete optimization workflow. The process begins with Sobol initialization to provide broad coverage of the design space, then transitions to acquisition-guided sampling. At each iteration, the selected design is evaluated through closed-loop Chrono::CRM simulation, and the result is used to update the GP surrogate before selecting the next candidate.

%% file: sections/results.tex
\section{Results}
\label{sec:results}

\subsection{Wheel-Only Optimization}
\label{subsec:wheel_only}
As an initial study, we optimize wheel geometry alone while keeping control parameters fixed. This serves as a baseline before introducing coupled control design. The design vector is restricted to the five wheel parameters defined in Sec.~\ref{subsec:bayesian_optimization}:
\[
\mathbf{x} = [r_o, w_r, g_r, n_g, \alpha_g]^\top \in \mathbb{R}^5.
\]
Performance is evaluated using a straight-line pull test.

\subsubsection{Setup}
\begin{figure}[H]
    \centering
    \includegraphics[width=1.0\textwidth]{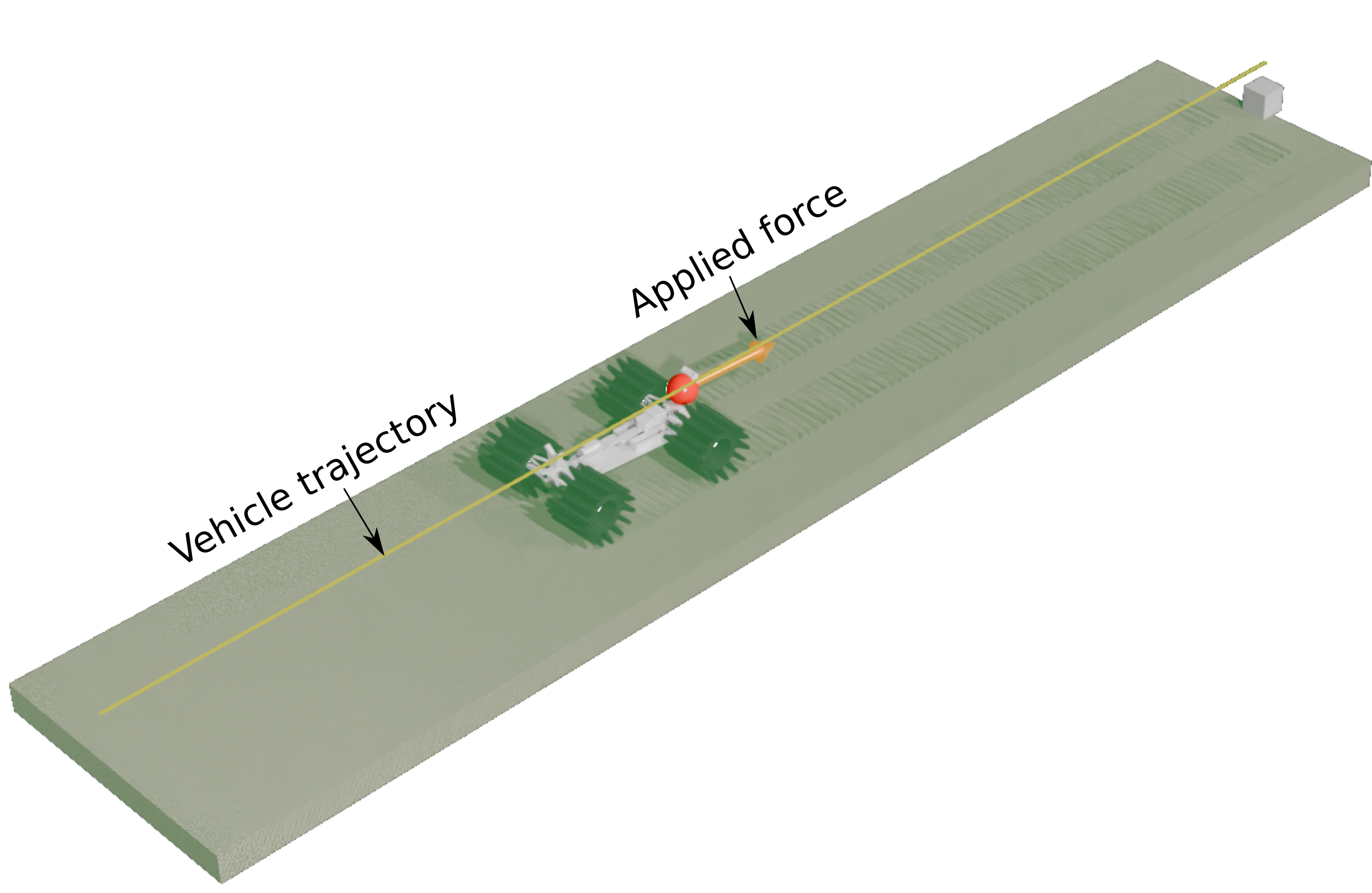}
    \caption{Pull test configuration. Yellow: vehicle trajectory; orange arrow: applied resistive force direction; red dot: force application point; gray block: fixed anchor.}
    \label{fig:pull_setup}
\end{figure}

The pull test simulates a vehicle traversing deformable terrain while towing a constant load, representative of high drawbar demand scenarios. The terrain consists of a $5\,\si{m} \times 1\,\si{m} \times 0.1\,\si{m}$ patch of loose sand ($\rho = 1700\,\si{kg/m^3}$, $\mu_s = \mu_2 = 0.6$, zero cohesion). The vehicle moves in a straight line at $70\%$ throttle while a constant $25\,\si{N}$ horizontal force is applied at the rear to emulate drawbar load (Fig.~\ref{fig:pull_setup}).

The terrain and wheel geometry are discretized with SPH particle spacing of \SI{0.5}{cm}, time step \SI{2e-4}{s} (CFL-based), and kernel radius $1.2$ times the particle spacing~\cite{Huzaifa2025CRM}. Only traversal speed is optimized ($w_s = 1.0$, $w_t = w_p = 0$). The optimization uses $n_{\text{sobol}} = 1200$ Sobol samples followed by $800$ acquisitions via Log Noisy Expected Improvement, totaling $T = 2000$ evaluations.

\subsubsection{Sensitivity Analysis}
We perform global sensitivity analysis using Sobol indices to assess parameter importance. Since direct computation on the high-fidelity model is prohibitive, we train a surrogate model on all 2000 designs and estimate first-order ($S_1$) and total-order ($S_T$) indices via the Saltelli scheme with 40,000 samples (Table~\ref{tab:pull_sens}). Among three candidate surrogates-- Gaussian Process, XGBoost, and Random Forest -- Random Forest achieved the highest predictive accuracy under five-fold cross-validation ($R^2 = 0.93$) and is therefore used for all sensitivity analyses reported in this work.

\begin{table}[!htbp]
    \centering
    \caption{Sensitivity indices for wheel-only optimization (pull test). $S_1$: main effects; $S_T$: main effects plus interactions; PDP Swing: marginal effect magnitude from ICE analysis.}
    \label{tab:pull_sens}
    \begin{tabular}{l c c c}
        \hline
        Parameter & $S_1$ & $S_T$ & PDP Swing (\%) \\
        \hline
        $r_o$ & 0.773 & 0.868 & 18.65 \\
        $w_r$ & 0.067 & 0.175 & 6.35 \\
        $n_g$ & 0.060 & 0.079 & 1.57 \\
        $g_r$ & 0.051 & 0.162 & 1.31 \\
        $\alpha_g$ & 0.022 & 0.026 & 1.7 \\
        \hline
    \end{tabular}
\end{table}

The outer radius $r_o$ is the dominant contributor to objective variance, accounting for approximately 77\% of the main effect and nearly 87\% when interactions are included. This indicates that overall wheel size primarily governs performance in the pull test. Secondary contributions arise from the width ratio $w_r$ and grouser-related parameters ($n_g$, $g_r$), with total-order indices indicating non-negligible interaction effects, particularly for $w_r$ and $g_r$. In contrast, the grouser angle $\alpha_g$ exhibits minimal influence over the explored design range.

\begin{figure}[H]
    \centering
    \includegraphics[width=1.0\textwidth]{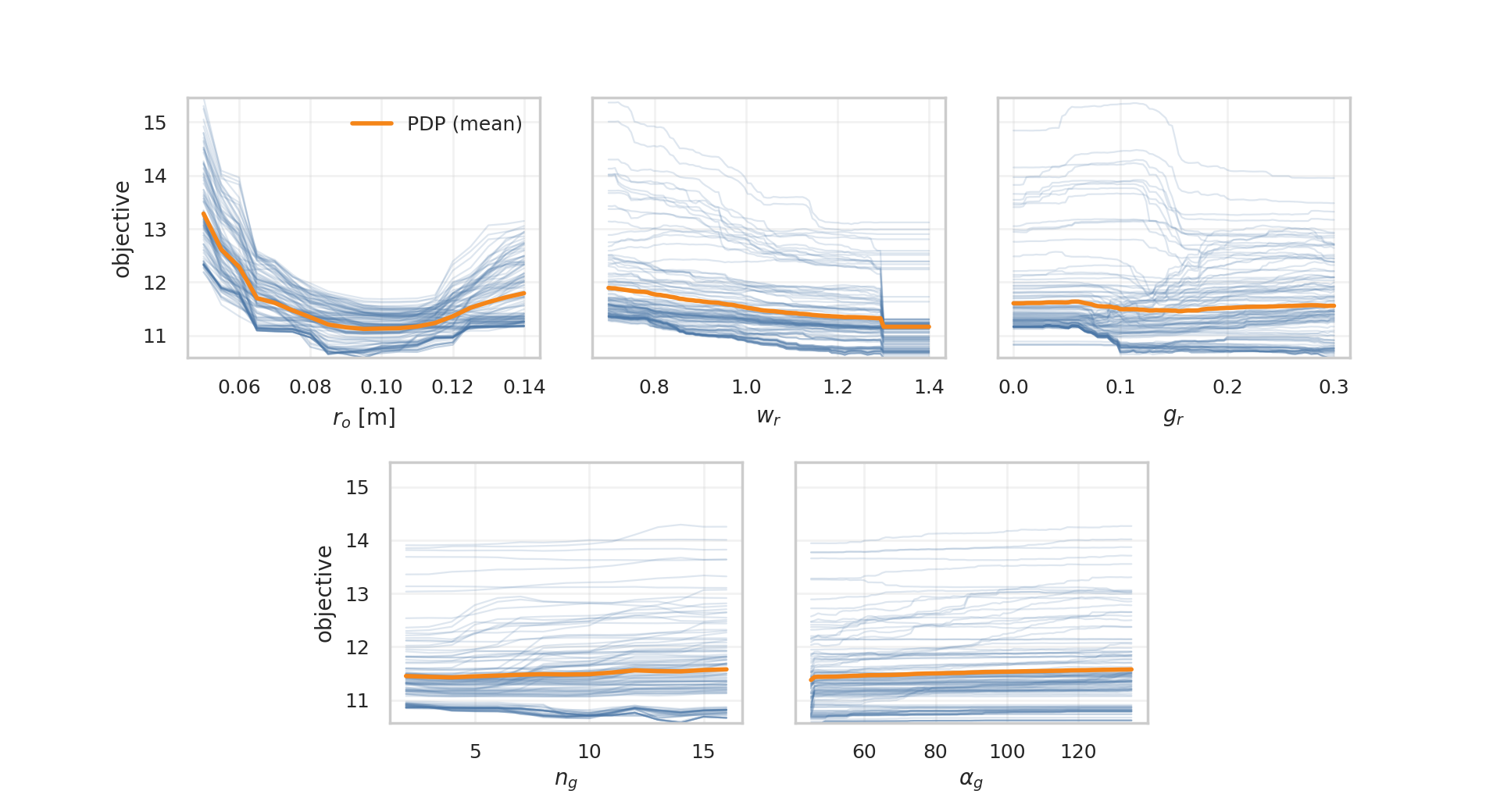}
    \caption{ICE plots for wheel parameters (pull test). Each curve represents one conditioning point; bold line is the Partial Dependence Plot (mean trend). Based on 120 conditioning points and 100 grid locations per sweep.}
    \label{fig:pull_ice}
\end{figure}

While Sobol indices quantify parameter importance in terms of variance contribution, they do not indicate the directionality of each parameter's effect. To examine how individual parameters influence the objective, we analyze the surrogate model using Individual Conditional Expectation (ICE) plots (Fig.~\ref{fig:pull_ice}). These plots show the predicted objective as a single parameter is varied while holding all others fixed at observed values, with the corresponding Partial Dependence Plot (PDP) representing the mean trend across all conditioning points. Consistent with the Sobol analysis, the outer radius $r_o$ has the strongest marginal effect, with a clear optimum in the range of approximately \SI{0.09}{m} to \SI{0.12}{m}; in other words, bigger is not necessarily better. The wheel width exhibits a weaker but consistent preference toward larger values, while the effects of grouser height, number, and angle are comparatively minor, as confirmed by the PDP swing values reported in Table~\ref{tab:pull_sens}.

\subsubsection{Experimental Validation}
To evaluate the transferability of simulation-optimized designs to physical hardware, we select three wheels from the BO results: the best-performing wheel, the 100th-ranked wheel, and the 1000th-ranked wheel. The geometric specifications of these wheels are listed in Table~\ref{tab:pull_wheels}. The corresponding CAD models are extracted from the optimized parameters and fabricated using a 3D printer.

\begin{table}[h]
    \centering
    \caption{Wheel specifications for experimental validation.}
    \label{tab:pull_wheels}
    \begin{tabular}{c c c c c c}
        \toprule
        Rank & $r_o$ (cm) & $w_r$ & $g_r$ & $n_g$ & $\alpha_g$ (deg) \\
        \midrule
        1 & 9.5 & 1.3 & 0.3 & 14 & 45 \\
        100 & 8 & 0.81 & 0.06 & 12 & 113 \\
        1000 & 5 & 0.7 & 0.1 & 4 & 112 \\
        \bottomrule
    \end{tabular}
\end{table}

\begin{figure}[H]
    \centering
    \includegraphics[width=1.0\textwidth]{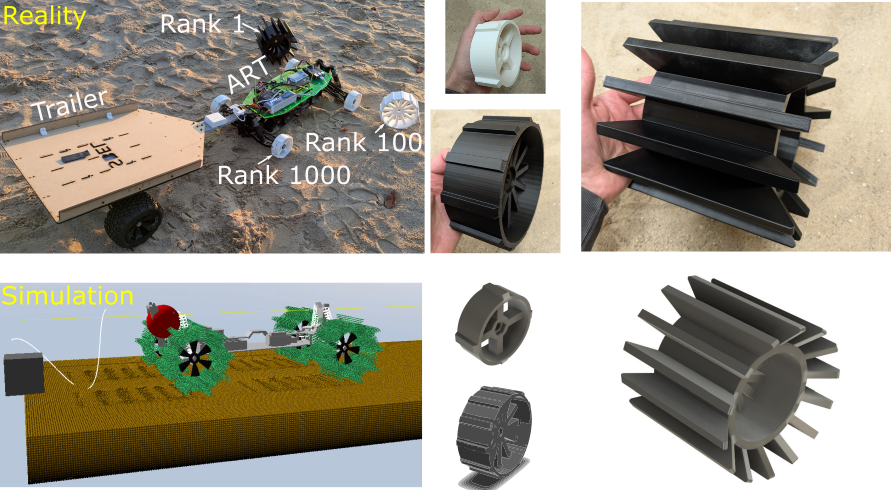}
    \caption{Experimental validation setup: physical \texttt{ART} vehicle with 3D-printed wheels (top) and corresponding simulation configurations (bottom). Image also shows a close-up of the three wheels employed in the virtual and physical tests.}
    \label{fig:real_vs_sim}
\end{figure}

Physical validation is conducted using the \texttt{ART} vehicle pulling a trailer outdoors on loose sand. The trailer is equipped with rubber wheels and has a total mass of \SI{3}{kg}. A comparison between the simulation and physical test setups is shown in Fig.~\ref{fig:real_vs_sim}. Unlike the simulation, which applies a constant horizontal drawbar force of \SI{25}{N}, the physical experiments involve an uncalibrated towing configuration and approximate soil properties. As a result, absolute traversal times are not expected to match between simulation and experiment. Instead, the objective of this test is to assess whether relative performance trends predicted in simulation are preserved in physical tests.

Table~\ref{tab:real_vs_sim} shows that while physical traversal times are longer than simulated, the performance ranking is preserved: the simulation-optimal wheel performs best in hardware, followed by the 100th- and 1000th-ranked designs. Although material properties differ between environments, the simulation's value lies in its predictive power; it successfully identifies the highest-performing physical designs, validating the optimization's qualitative accuracy.

\begin{table}[h]
    \centering
    \caption{Simulation vs.\ physical traversal times. Rankings are preserved despite absolute time differences.}
    \label{tab:real_vs_sim}
    \begin{tabular}{ccc}
      \toprule
      Rank & Sim.\ Time (s) & Physical Time (s) \\
      \midrule
      1 & 2.31 & 8.0 \\
      100 & 2.63 & 10.3 \\
      1000 & 3.20 & 15.0 \\
      \bottomrule
    \end{tabular}
\end{table}

\subsection{Co-Design of Wheel and Controller Optimization}
\label{subsec:joint_opt}
This second study extends the framework to simultaneously optimize the wheel geometry and the steering PID gains. The design vector now includes eight parameters:
\[
\mathbf{x} = [r_o, w_r, g_r, n_g, \alpha_g, K_{p,s}, K_{i,s}, K_{d,s}]^\top \in \mathbb{R}^8.
\]
Since steering gains cannot be identified from straight-line motion, we employ a sinusoidal path-following test that excites lateral dynamics while evaluating traction and tracking performance.

\subsubsection{Setup}
The sine curve test requires the vehicle to follow a sinusoidal trajectory on deformable terrain while subjected to a constant resistive load, thereby coupling wheel--terrain interaction with steering control performance. The terrain configuration matches that of the pull test, consisting of a $5.0\,\si{m} \times 1.0\,\si{m}$ patch with a depth of $0.10\,\si{m}$, modeled as loose sand with the same material parameters described in Sec.~\ref{subsec:wheel_only}.

\begin{figure}[H]
    \centering
    \includegraphics[width=1.0\textwidth]{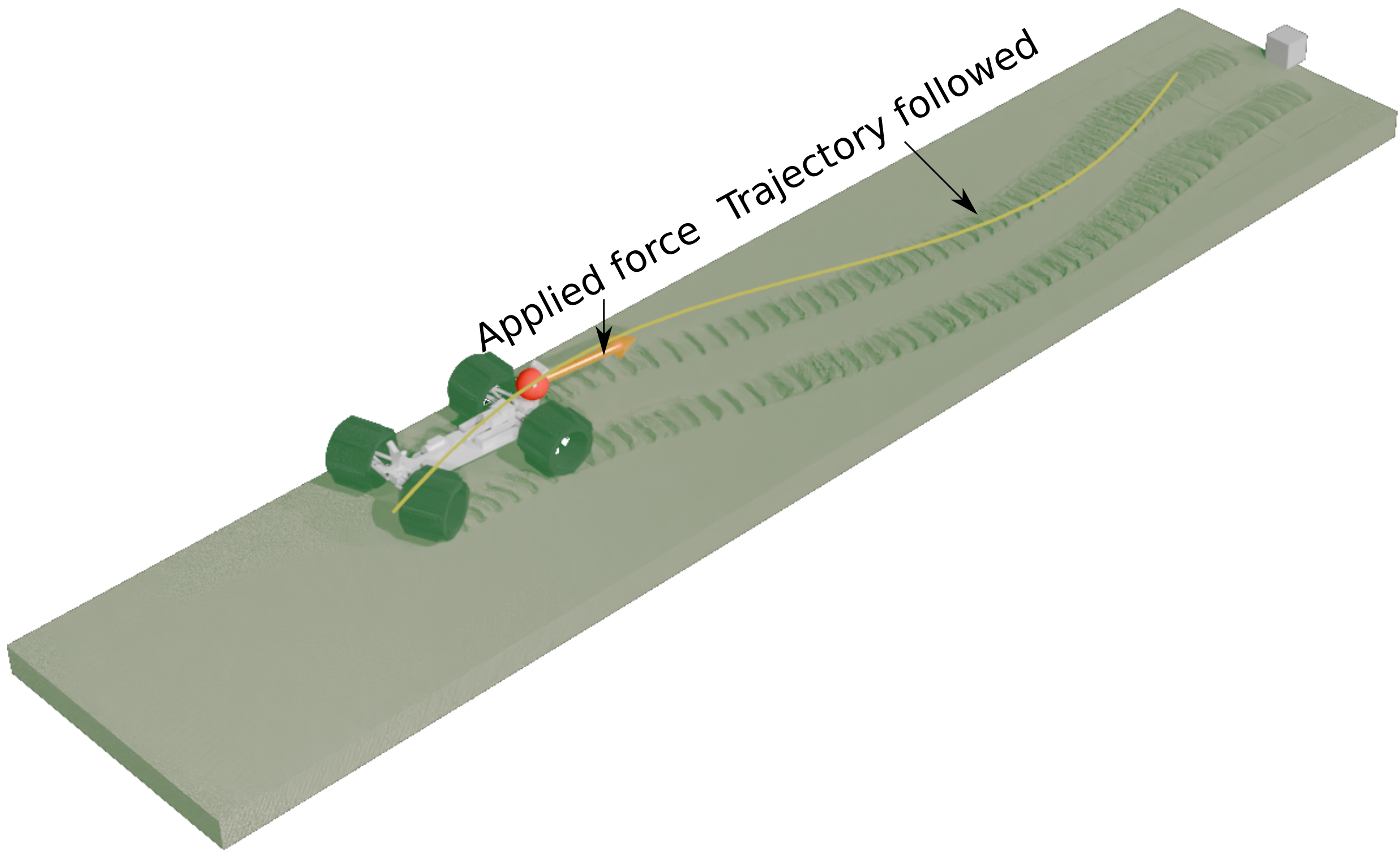}
    \caption{Sine test configuration. Visual elements follow the same convention as Fig.~\ref{fig:pull_setup}.}
    \label{fig:sine_setup}
\end{figure}

The reference trajectory is defined as a Bezier curve interpolating waypoints sampled from a sinusoidal function with a lateral amplitude of $0.2\,\si{m}$ and two full periods over the length of the terrain. A path-following driver guides the vehicle along this trajectory using a steering PID controller with a look-ahead distance of $0.5\,\si{m}$, while a separate PID speed controller regulates throttle to maintain a target forward velocity of $3\,\si{m/s}$. As in the pull test, a constant horizontal load of $25\,\si{N}$ is applied at the rear of the vehicle. As the vehicle rotates, the force rotates to connect to the anchor points (gray box). A schematic of the sine-test configuration is shown in Fig.~\ref{fig:sine_setup}.

SPH parameters match those in Sec.~\ref{subsec:wheel_only}. Performance metrics include RMS cross-track error, traversal time, and average mechanical power.

\subsubsection{Optimization Strategies}
We compare two strategies under equivalent computational budgets ($T = 3000$ total simulations):

\begin{enumerate}
    \item \textbf{Joint Optimization (Simultaneous Co-Design):} In the first strategy, wheel geometry and steering PID controller gains are optimized simultaneously within a single Bayesian Optimization loop. The objective function weights are set to $w_s = 0.4$, $w_t = 0.2$, and $w_p = 0.4$, placing equal emphasis on traversal speed and energy consumption, with a smaller but nonzero weight on path-tracking accuracy. The optimization is initialized with $n_{\text{sobol}} = 1800$ Sobol samples, followed by $1200$ samples selected using the Log Noisy Expected Improvement acquisition function, resulting in a total of $T = 3000$ evaluated designs.

    \item \textbf{Looped Optimization (Sequential Decoupled Design):} In the second strategy, wheel geometry and steering controller gains are optimized sequentially in two stages. First, wheel geometry is optimized on the sine test while holding the steering PID gains fixed at $K_{p,s} = 5.5$, $K_{i,s} = 0$, and $K_{d,s} = 1$. In this stage, the objective function weights are set to $w_s = 0.5$, $w_t = 0$, and $w_p = 0.5$, focusing exclusively on speed and energy efficiency. This wheel-only stage uses $n_{\text{sobol}} = 1200$ Sobol samples followed by $800$ Bayesian Optimization samples, for a total of $T = 2000$ evaluations.

    In the second stage, the optimized wheel geometry from the first stage is held fixed, and the steering PID gains are optimized using the same sine test. Here, the objective weights are set to $w_s = 0$, $w_t = 1$, and $w_p = 0$, isolating path-tracking performance. This stage uses $n_{\text{sobol}} = 600$ initial samples and $400$ BO samples, yielding $T = 1000$ additional evaluations. Across both stages, the looped optimization uses a total of $T = 3000$ simulations, matching the computational budget of the joint optimization strategy.
\end{enumerate}

\subsubsection{Sensitivity Analysis}
\label{subsubsec:joint_sensitivity}
We perform sensitivity analysis separately for each strategy to examine parameter identifiability.

\begin{figure}[H]
    \centering
    \includegraphics[width=1.0\textwidth]{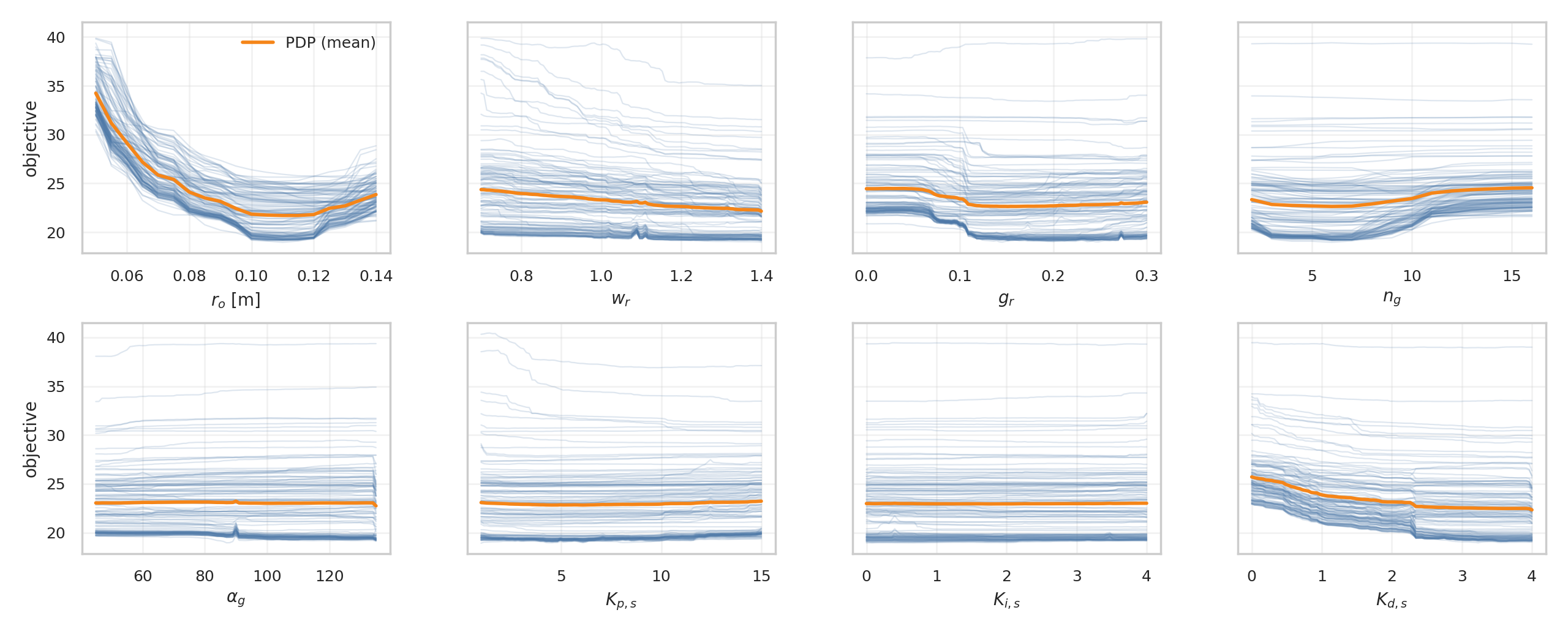}
    \caption{ICE plots for joint optimization. The flat PID curves reflect suppressed controller identifiability.}
    \label{fig:joint_ice}
\end{figure}

\paragraph{Joint Optimization}
Table~\ref{tab:joint_sens} and Fig.~\ref{fig:joint_ice} show sensitivity results for the joint case (RF surrogate $R^2 = 0.926$). The outer radius again dominates, accounting for 82\% of first-order and 88\% of total variance, even stronger than in the pull test. Steering gains contribute minimally: $K_{d,s}$ accounts for only 2.4\% of variance, while $K_{p,s}$ and $K_{i,s}$ are negligible. This occurs because tracking error constitutes only 20\% of the objective, limiting the influence of steering parameters.

\begin{table}[!htbp]
    \centering
    \caption{Sensitivity indices for joint optimization (sine test).}
    \label{tab:joint_sens}
    \begin{tabular}{l c c c}
        \hline
        Parameter & $S_1$ & $S_T$ & PDP Swing (\%) \\
        \hline
        $r_o$ & 0.828 & 0.884 & 50.96 \\
        $w_r$ & 0.049 & 0.090 & 9.54 \\
        $K_{d,s}$ & 0.024 & 0.044 & 14.42 \\
        $g_r$ & 0.022 & 0.042 & 7.94 \\
        $n_g$ & 0.008 & 0.023 & 8.13 \\
        $\alpha_g$ & 0.005 & 0.006 & 2.09 \\
        $K_{p,s}$ & 0.003 & 0.007 & 1.74 \\
        $K_{i,s}$ & 0.001 & 0.001 & 0.31 \\
        \hline
    \end{tabular}
\end{table}

\begin{figure}[H]
    \centering
    \includegraphics[width=1.0\textwidth]{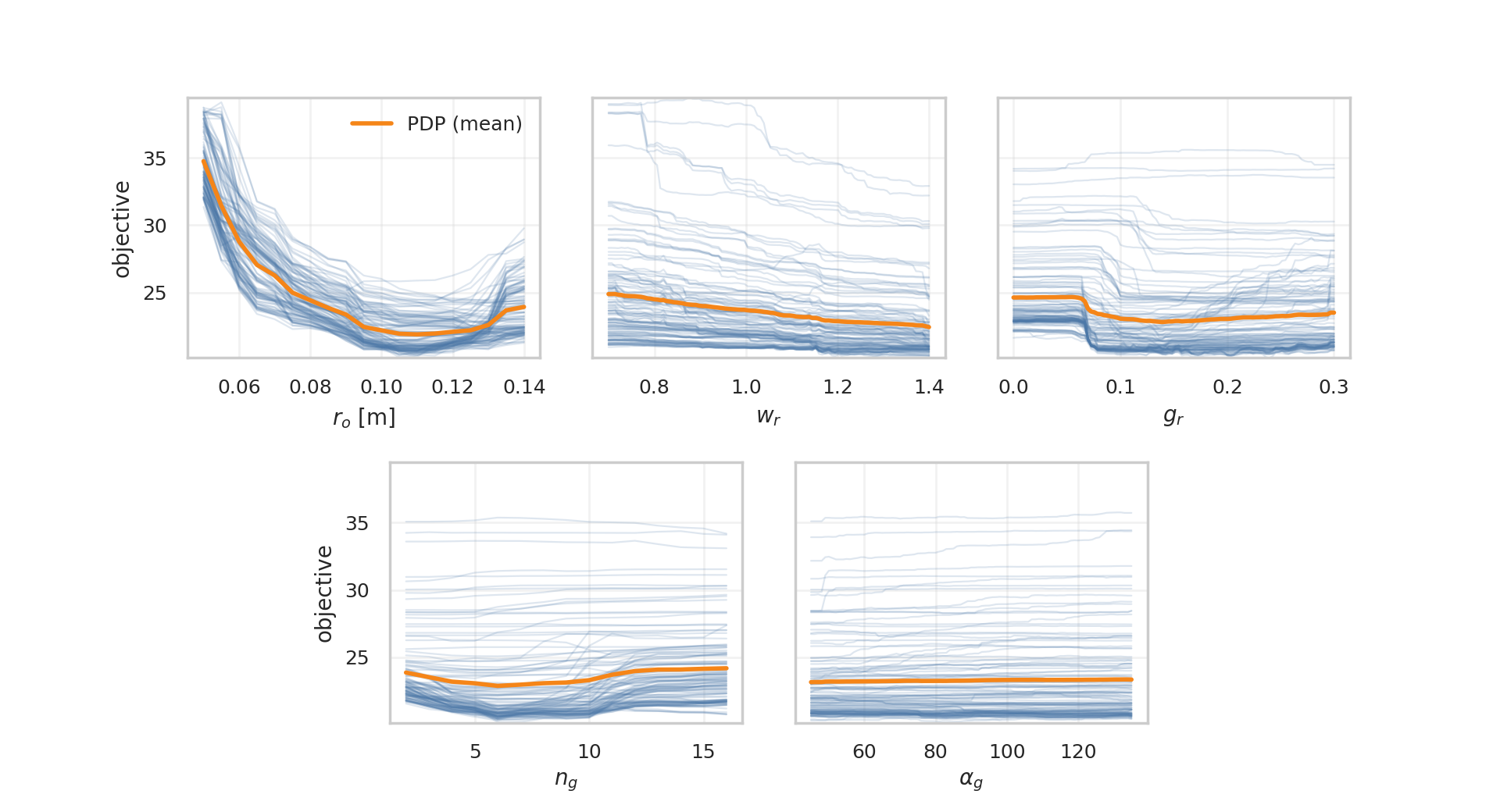}
    \caption{ICE plots for looped optimization, Stage 1 (wheel parameters).}
    \label{fig:looped_wheel_ice}
\end{figure}

\paragraph{Looped Optimization: Wheel Phase}
In Stage 1, wheel geometry is optimized with fixed controller gains (RF surrogate $R^2 = 0.947$). Results (Table~\ref{tab:looped_wheel_sens}, Fig.~\ref{fig:looped_wheel_ice}) are similar to the joint case: $r_o$ dominates with $S_1 = 0.825$.

\begin{table}[!htbp]
    \centering
    \caption{Sensitivity indices for looped optimization, Stage 1 (wheel only, fixed PID).}
    \label{tab:looped_wheel_sens}
    \begin{tabular}{l c c c}
        \hline
        Parameter & $S_1$ & $S_T$ & PDP Swing (\%) \\
        \hline
        $r_o$ & 0.825 & 0.899 & 52.06 \\
        $w_r$ & 0.056 & 0.096 & 10.46 \\
        $n_g$ & 0.019 & 0.028 & 5.56 \\
        $g_r$ & 0.013 & 0.066 & 7.91 \\
        $\alpha_g$ & 0.004 & 0.005 & 0.86 \\
        \hline
    \end{tabular}
\end{table}

\begin{figure}[H]
    \centering
    \includegraphics[width=1.0\textwidth]{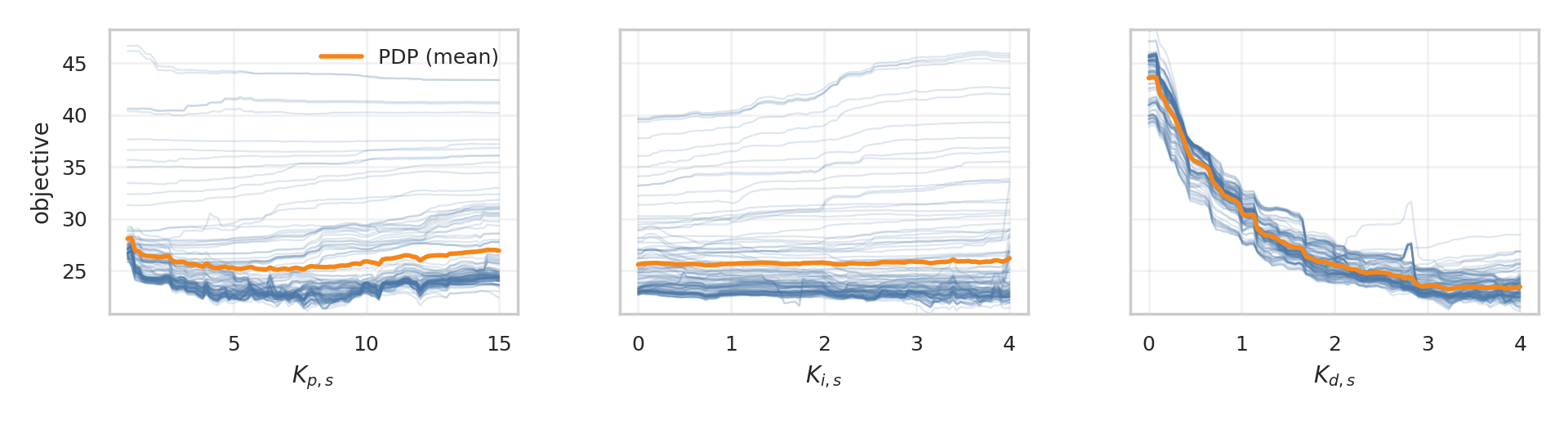}
    \caption{ICE plots for looped optimization, Stage 2 (PID parameters).}
    \label{fig:looped_pid_ice}
\end{figure}

\paragraph{Looped Optimization: Controller Phase}
In Stage 2, wheel geometry is fixed and only PID gains are optimized with a tracking-focused objective ($w_t = 1$). The RF surrogate achieves $R^2 = 0.929$. Results (Table~\ref{tab:looped_pid_sens}, Fig.~\ref{fig:looped_pid_ice}) show that $K_{d,s}$ now accounts for 91\% of first-order variance, demonstrating strong controller identifiability when the objective isolates tracking performance.

\begin{table}[!htbp]
    \centering
    \caption{Sensitivity indices for looped optimization, Stage 2 (PID only, fixed wheel).}
    \label{tab:looped_pid_sens}
    \begin{tabular}{l c c c}
        \hline
        Parameter & $S_1$ & $S_T$ & PDP Swing (\%) \\
        \hline
        $K_{d,s}$ & 0.910 & 0.975 & 72.66 \\
        $K_{p,s}$ & 0.016 & 0.039 & 11.79 \\
        $K_{i,s}$ & 0.005 & 0.032 & 2.67 \\
        \hline
    \end{tabular}
\end{table}

\paragraph{Comparison and Interpretation}
The two optimization strategies differ in how they handle parameter identifiability. In joint optimization, wheel geometry dominates the composite objective, which suppresses the influence of steering gains. When all parameters compete within a single objective, the optimizer naturally favors those with the largest global effect, in this case, the wheel radius. The looped strategy avoids this by aligning each stage with parameters that directly affect the corresponding objective component, allowing both wheel and controller sensitivities to be resolved.

From a practical standpoint, joint optimization captures coupling effects between mechanical design and control, while sequential optimization offers improved interpretability. It is also possible to iterate the looped process by re-optimizing wheel geometry with the tuned controller gains. We discuss this in more detail in Sec.~\ref{sec:future}.

\subsubsection{Performance Comparison}
Table~\ref{tab:optimized_wheels_and_controllers} lists the optimized wheel geometry and steering PID controller gains obtained using the joint and looped optimization approaches. Table~\ref{tab:metrics_comparison} compares the corresponding traversal time, average power consumption, and RMS tracking error achieved on the sine curve test.

\begin{table}[h]
    \centering
    \caption{Optimized wheel geometry and controller gains.}
    \label{tab:optimized_wheels_and_controllers}
    \begin{tabular}{c c c c c c c c c}
        \toprule
        Approach & $r_o$ (cm) & $w_r$ & $g_r$ & $n_g$ & $\alpha_g$ (deg) & $K_{p,s}$ & $K_{i,s}$ & $K_{d,s}$ \\
        \midrule
        Joint & 10.5 & 1.4 & 0.21 & 7 & 135 & 6.11 & 4.0 & 4.0 \\
        Looped & 11 & 1.4 & 0.24 & 9 & 135 & 12.1 & 2.2 & 3.82 \\
        \bottomrule
    \end{tabular}
\end{table}

\begin{table}[h]
    \centering
    \caption{Performance on sine test.}
    \label{tab:metrics_comparison}
    \begin{tabular}{l c c c}
        \toprule
        Approach & Time (s) & Power (W) & RMS Error (m) \\
        \midrule
        Joint & 2.39 & 205.38 & 0.068 \\
        Looped (default controller) & 2.40 & 207.47 & 0.082 \\
        Looped (optimized controller) & 2.41 & 206.01 & 0.072 \\
        \bottomrule
    \end{tabular}
\end{table}

Both optimization strategies converge to similar wheel geometries. In particular, the outer radius, width ratio, and grouser angle are nearly identical between the two solutions, suggesting that the optimal wheel design for the sine test is largely insensitive to whether controller gains are optimized jointly or sequentially. Differences in the remaining wheel parameters are modest and do not substantially affect overall performance.

In contrast, the optimized steering controller gains differ between the two approaches, reflecting the different treatment of controller tuning in the optimization process. In the joint case, steering gains are selected in the presence of competing objectives related to speed and energy consumption, whereas in the looped approach the controller gains are optimized separately with a tracking-focused objective.

Despite these differences in controller gains, traversal time and power consumption are nearly identical across all cases. The RMS tracking error shows differences: the joint optimization yields the lowest tracking error (0.068~m), while the looped approach exhibits a reduction in tracking error from 0.082~m to 0.072~m after the controller is optimized in the second phase. Overall, both approaches achieve comparable performance, with small differences in tracking behavior attributable to how steering control is incorporated into the optimization.

\subsection{Generalization to Unseen Trajectory}
\label{subsec:generalization}
To assess whether optimized designs generalize beyond the training trajectory, we evaluate performance on a previously unseen racetrack.

\subsubsection{Racetrack Setup}
\begin{figure}[H]
    \centering
    \begin{subfigure}[b]{1.0\textwidth}
        \centering
        \includegraphics[width=1.0\textwidth]{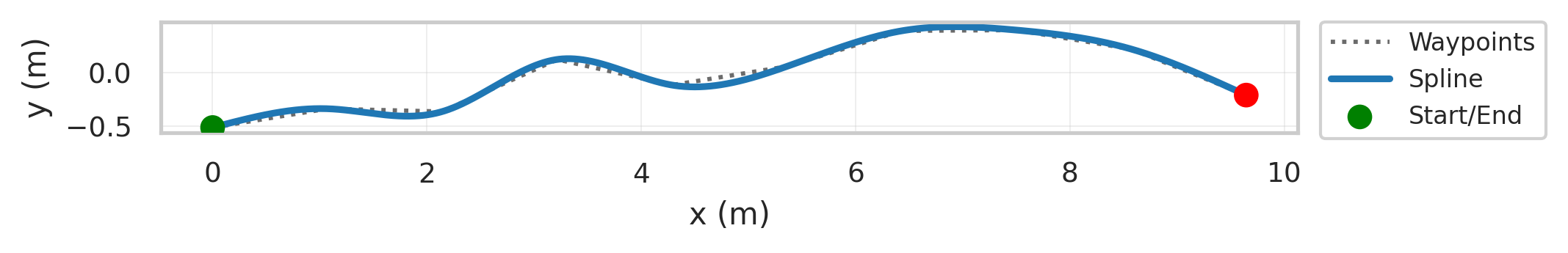}
        \caption{Waypoints and interpolated spline centerline.}
        \label{fig:racetrack}
    \end{subfigure}
    \vspace{0.5cm}
    \begin{subfigure}[b]{1.0\textwidth}
        \centering
        \includegraphics[width=1.0\textwidth]{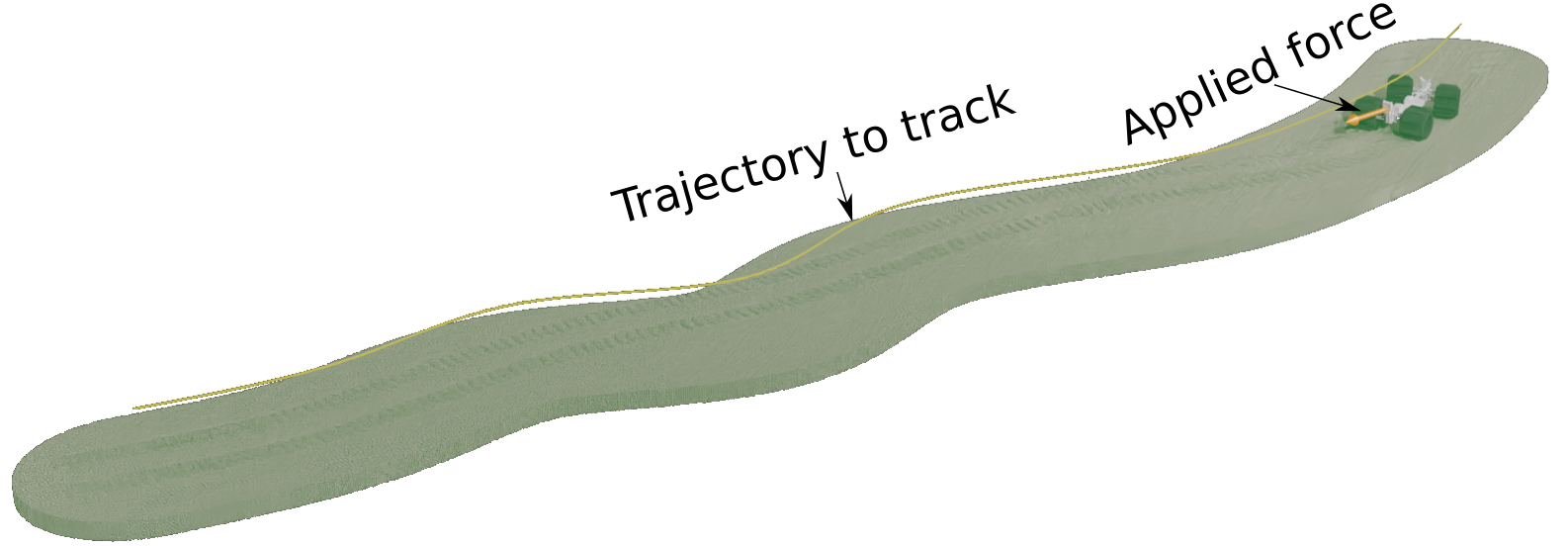}
        \caption{Simulation configuration. Visual conventions follow Fig.~\ref{fig:pull_setup}.}
        \label{fig:racetrack_terrain}
    \end{subfigure}
    \caption{Racetrack configuration for generalization testing. The trajectory is generated by fitting a spline through ten randomly placed waypoints, rescaled to \SI{10}{m} total length with a minimum turning radius of \SI{0.6}{m}. Deformable terrain fills a \SI{1.2}{m}-wide corridor around the centerline.}
    \label{fig:racetrack_combined}
\end{figure}

A randomized racetrack is generated by fitting a spline through ten control waypoints, rescaled to \SI{10}{m} total length with minimum turning radius \SI{0.6}{m}. Deformable terrain fills a \SI{1.2}{m}-wide corridor around the centerline, using the same material parameters as previous tests. The path-following controller uses the same look-ahead distance (\SI{0.5}{m}) and resistive load (\SI{25}{N}) as the sine test (Fig.~\ref{fig:racetrack_combined}).

\subsubsection{Results}
Table~\ref{tab:racetrack_results} summarizes racetrack performance. Performance rankings from the sine test are preserved: joint optimization and looped (with optimized controller) achieve identical traversal times (3.95~s), while looped with default controller is slower (4.02~s).

\begin{table}[h]
    \centering
    \caption{Racetrack performance (unseen trajectory).}
    \label{tab:racetrack_results}
    \begin{tabular}{l c c c}
        \toprule
        Approach & Time (s) & RMS Error (m) & Power (W) \\
        \midrule
        Joint & 3.95 & 0.080 & 242.05 \\
        Looped (default controller) & 4.02 & 0.140 & 240.35 \\
        Looped (optimized controller) & 3.95 & 0.093 & 247.47 \\
        \bottomrule
    \end{tabular}
\end{table}

The most notable difference is in tracking accuracy. RMS errors increase from 0.068--0.082~m (sine test) to 0.080--0.140~m (racetrack), reflecting the more demanding trajectory. The looped approach with default controller exhibits the worst tracking (0.140~m), but optimizing the controller reduces error by 34\% to 0.093~m. Joint optimization achieves the lowest error (0.080~m). This also shows that both optimization strategies produce designs capable of accurate path following on trajectories outside the training distribution.

\subsection{Computational Cost Analysis}
\label{subsec:comp_cost}

\subsubsection{Wheel-Only Optimization}
\begin{figure}[H]
    \centering
    \includegraphics[width=0.5\textwidth]{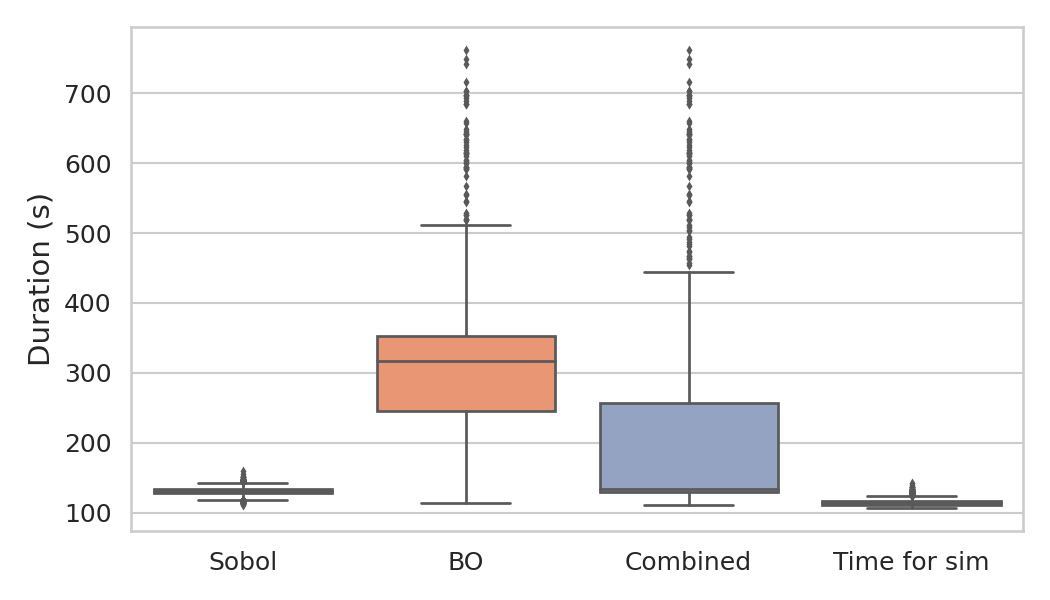}
    \caption{Computational time per sample (pull test).}
    \label{fig:pull_comp_time}
\end{figure}

Figure~\ref{fig:pull_comp_time} shows the distribution of computational time per sample for the pull test optimization. Three timing categories are distinguished: (i) the CRM simulation alone, which evaluates a single wheel design; (ii) Sobol samples, which run the simulation plus minor overhead for logging and data management; and (iii) Bayesian Optimization samples, which run the simulation and additionally retrain the Gaussian Process surrogate model after each evaluation. The CRM simulation alone requires approximately 114~s on average. Sobol samples require approximately 131~s on average, reflecting minimal overhead beyond the simulation itself. Bayesian Optimization samples require approximately 310~s on average; the additional time is due to refitting the GP surrogate to the growing dataset after each new observation. The total time for 1200 Sobol samples is approximately 44 hours (1.8 days), and for 800 Bayesian Optimization samples approximately 69 hours (2.9 days), yielding a total optimization time of approximately 113 hours (4.7 days).

\subsubsection{Looped Optimization}
\begin{figure}[H]
    \centering
    \begin{subfigure}[b]{0.48\textwidth}
        \centering
        \includegraphics[width=\textwidth]{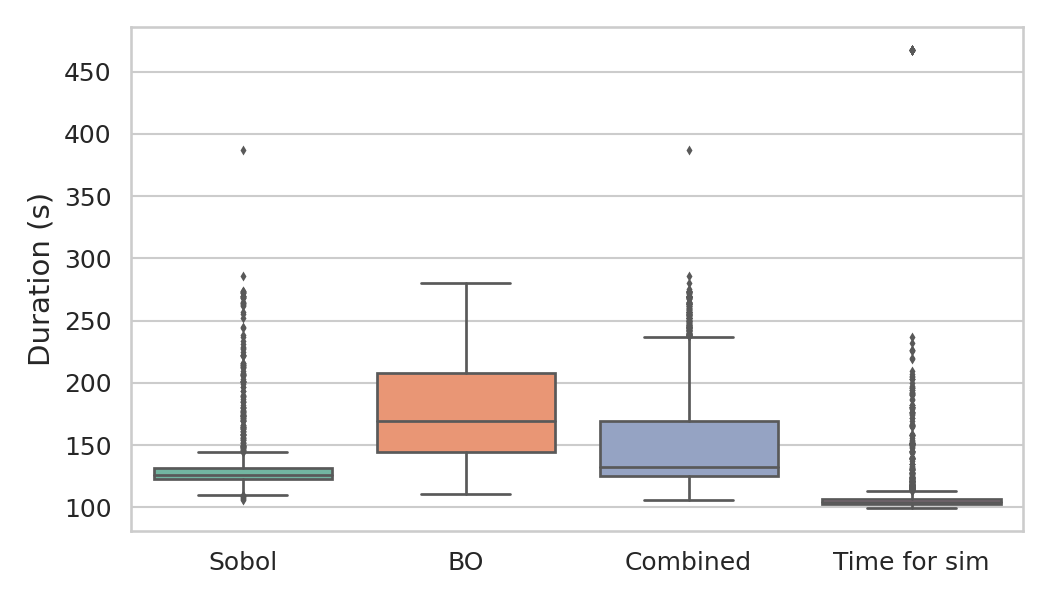}
        \caption{Phase 1 (wheel optimization)}
        \label{fig:loopedPhase1_comp_time}
    \end{subfigure}
    \hfill
    \begin{subfigure}[b]{0.48\textwidth}
        \centering
        \includegraphics[width=\textwidth]{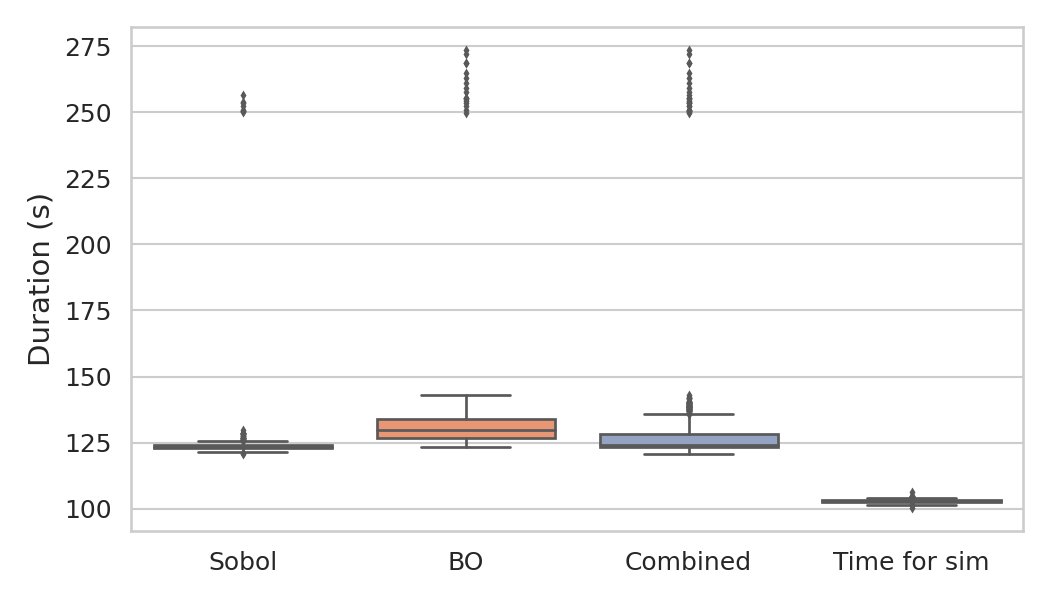}
        \caption{Phase 2 (controller optimization)}
        \label{fig:loopedPhase2_comp_time}
    \end{subfigure}
    \caption{Computational time per sample (looped optimization).}
    \label{fig:looped_comp_time}
\end{figure}

Figure~\ref{fig:looped_comp_time} shows the distribution of computational time per sample for both phases of the looped optimization. In Phase~1 (wheel optimization), the total time for 1200 Sobol samples is approximately 45~hours (1.9~days), and for 800 Bayesian Optimization samples approximately 39~hours (1.6~days). In Phase~2 (controller optimization), the total time for 600 Sobol samples is approximately 20~hours (0.8~days), and for 400 Bayesian Optimization samples approximately 14~hours (0.6~days). The overall computational cost for the looped optimization is approximately 118~hours (4.9~days).

\subsubsection{Joint Optimization}
\begin{figure}[H]
    \centering
    \includegraphics[width=0.5\textwidth]{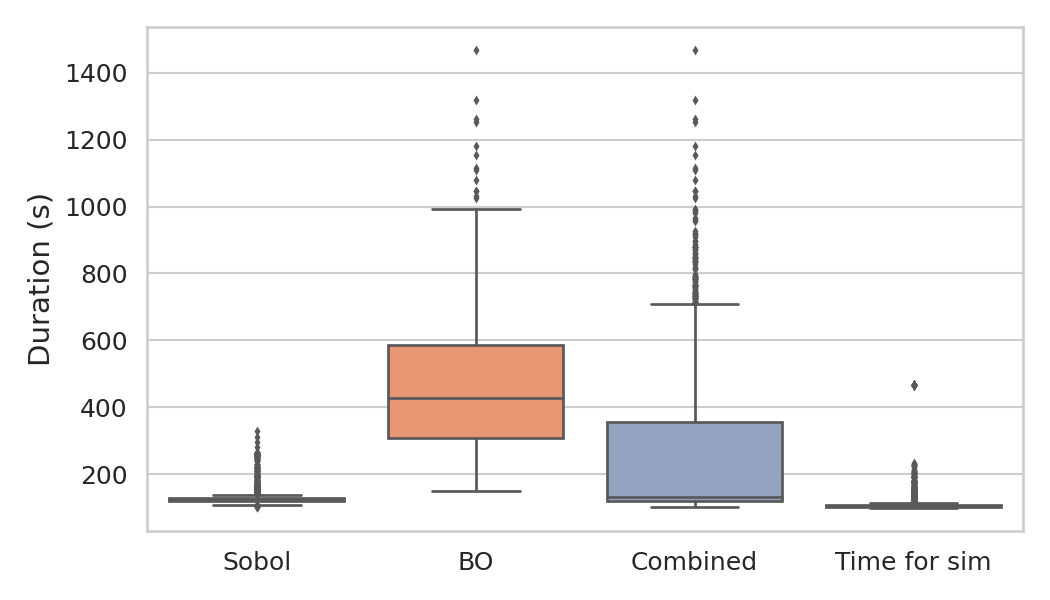}
    \caption{Computational time per sample (joint optimization).}
    \label{fig:joint_comp_time}
\end{figure}

Figure~\ref{fig:joint_comp_time} shows the distribution of computational time per sample for the joint optimization. The CRM simulation alone requires approximately 100~s on average. However, due to the overhead of retraining the Gaussian Process surrogate model after each evaluation, Bayesian Optimization samples require approximately 400~s on average. This overhead is larger than in the wheel-only case because the joint optimization operates in an eight-dimensional design space, making the GP surrogate more expensive to fit and query. For the joint optimization, the total time for 1800 Sobol samples is approximately 65 hours (2.7 days), and for 1200 Bayesian Optimization samples approximately 147 hours (6.15 days), yielding a total of 212 hours (8.8 days).

\subsection{Summary}
Joint optimization achieves slightly better tracking accuracy on both the sine test (RMS error of 0.068~m vs 0.072~m) and the racetrack (0.080~m vs 0.093~m), but requires 212 hours compared to 118 hours for the looped approach. The looped optimization reduces computational time by 44\% while achieving comparable traversal time and power consumption, with only a modest increase in tracking error. For applications where computational resources are limited or rapid design iteration is required, the looped strategy provides an efficient alternative to joint optimization.

%% file: sections/conclusion.tex
\section{Conclusion}
\label{sec:conclusion}

This work presented a full-vehicle, closed-loop co-optimization framework for rover wheel design and steering control, enabled by the computational efficiency of the continuum-based terramechanics simulator Chrono::CRM. Unlike prior DEM-based approaches that were constrained to single-wheel tests with prescribed motion, the proposed framework evaluates candidate wheel geometries and controller gains through complete vehicle--terrain simulations under realistic closed-loop operation, thereby capturing the coupling between mechanical design, control behavior, and deformable terrain interaction. Variance-based sensitivity analyses consistently identified the outer wheel radius as the dominant design parameter, accounting for approximately 77--88\% of objective variance across all test scenarios, while grouser geometry and orientation contributed secondary effects. It was noted that increasing the wheel radius improved performance up to a point, after which larger wheel radii had a detrimental effect. Preliminary experimental validation using 3D-printed wheels demonstrated that relative performance rankings predicted in simulation were preserved in physical hardware tests, despite uncalibrated terrain parameters and differences in loading configuration. Two optimization strategies were compared: simultaneous joint optimization of wheel and controller parameters, and a sequential looped approach that decouples mechanical and control design. Both strategies converged to similar wheel geometries and achieved comparable traversal times and energy consumption, though joint optimization yielded modestly lower tracking error on the training trajectory. The looped approach, however, reduced total computational time by approximately 44\% and restored parameter identifiability for the steering controller gains, which were otherwise suppressed when competing with wheel geometry in a composite objective. Evaluation on a previously unseen racetrack trajectory confirmed that optimized designs generalize beyond the training distribution, with tracking errors increasing moderately but performance orderings remaining consistent. From a computational perspective, the framework completed 3,000 full-vehicle deformable-terrain simulations within 4--9 days depending on optimization strategy, representing a substantial reduction compared to the weeks or months typically required by discrete element methods. These results suggest that continuum-based terramechanics simulation, combined with Bayesian Optimization, provides a practical pathway for systematic, vehicle-level exploration of rover wheel design spaces. The choice between joint and sequential optimization strategies presents a trade-off between capturing design-control coupling and achieving computational efficiency with improved interpretability.

%% file: sections/future.tex
\section{Future Work}
\label{sec:future}

Several directions remain for extending the framework presented in this work. A natural extension of the looped optimization strategy is to iterate the sequential process: after optimizing controller gains for a given wheel geometry, the wheel geometry could be re-optimized under the updated controller, and this alternation repeated until convergence. Such an iterative scheme may recover some of the coupling benefits of joint optimization while retaining the computational efficiency and interpretability of the sequential approach. Another avenue involves expanding the wheel parameterization to include more complex grouser geometries, such as chevron patterns or curved profiles, which prior DEM-based studies have shown to influence lateral traction and steering response. The experimental validation presented here employed uncalibrated terrain parameters; future work should incorporate systematic terrain calibration with Bayesian Inference~\cite{weiVirtualBevameter2024,huzaifa-calibrationJCND2023} to improve quantitative agreement between simulation and physical tests and enable more rigorous assessment of sim-to-real transfer fidelity. Extending the optimization to reduced-gravity conditions, relevant for lunar and Martian missions, would allow direct evaluation of wheel designs under planetary surface conditions where gravity-offset experiments have been shown to produce misleading results. Beyond these extensions, recent advances in multi-fidelity Bayesian optimization offer a promising path toward further reducing computational cost. By leveraging surrogate evaluations at coarser SPH resolutions or simplified terrain models, multi-fidelity schemes could accelerate design space exploration while reserving high-fidelity simulations for refinement near optima. Additionally, incorporating domain randomization over terrain parameters such as friction coefficient, cohesion, and bulk density during optimization could yield wheel designs that are robust across a range of soil conditions rather than tuned to a single terrain configuration. These directions collectively aim to improve the efficiency, generality, and physical fidelity of simulation-driven rover wheel design.